\documentclass[10pt,twocolumn,letterpaper]{article}

\usepackage[pagenumbers]{cvpr} % To force page numbers, e.g. for an arXiv version

\def\paperID{12064}
\def\confName{CVPR}
\def\confYear{2024}

\usepackage[dvipsnames]{xcolor}

\usepackage{multirow}
\usepackage[
  separate-uncertainty=true,
  multi-part-units=single,
  list-units=repeat,
  detect-weight,
  mode=text
]{siunitx}
\usepackage{nccmath}
\usepackage{bm}
\usepackage{xfrac}
\usepackage[accsupp]{axessibility}

\definecolor{cvprblue}{rgb}{0.21,0.49,0.74}
\usepackage[pagebackref,breaklinks,colorlinks,citecolor=cvprblue]{hyperref}

\DeclareMathOperator{\SO3}{\ensuremath{\mathbf{SO}(3)}}
\DeclareMathOperator{\SE3}{\ensuremath{\mathbf{SE}(3)}}
\DeclareMathOperator{\so3}{\ensuremath{\mathfrak{so}(3)}}
\DeclareMathOperator{\se3}{\ensuremath{\mathfrak{se}(3)}}

\def\successrate{87\%\xspace}
\def\bestmtre{\SI{0.9 \pm 2.8}{}\xspace}

\newcommand{\name}{DiffPose\xspace}
\newcommand{\subpara}[1]{\vspace{0.4em} \noindent \textbf{#1}}

\newcommand{\xray}{\mbox{X-ray}\xspace}
\newcommand{\xrays}{\mbox{X-rays}\xspace}

\title{Intraoperative 2D/3D Image Registration via Differentiable \xray Rendering}
\author{
    Vivek Gopalakrishnan$^{1,2}$
    \quad Neel Dey$^{2}$
    \quad Polina Golland$^{1,2}$ \\
    {\small $^{1}$Harvard-MIT Health Sciences and Technology \quad $^{2}$MIT Computer Science and Artificial Intelligence Laboratory} \\
    {\small \texttt{\{vivekg,dey,polina\}@csail.mit.edu}}
}

\begin{document}
\twocolumn[{
\renewcommand\twocolumn[1][]{#1}
\maketitle
\begin{center}
    \centering
    \captionsetup{type=figure}
    \includegraphics[width=\linewidth]{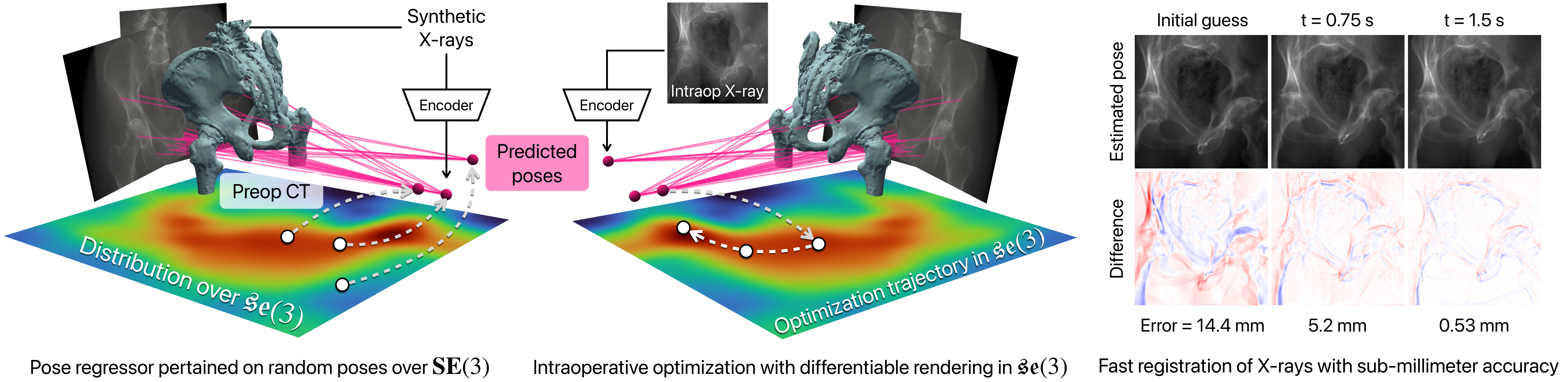}
    \caption{\textbf{We present \name, a self-supervised framework for differentiable 2D/3D registration.} Trained exclusively on synthetic \xrays rendered from a patient-specific preoperative CT scan, \name aligns intraoperative \xrays with sub-millimeter accuracy. \name does not require manually annotated training data, performs consistently across subjects, and registers images at clinically relevant speeds.}
    \label{fig:teaser}
\end{center}
}]

\begin{abstract}
Surgical decisions are informed by aligning rapid portable 2D intraoperative images (\eg \xrays) to a high-fidelity 3D preoperative reference scan (\eg CT). However, 2D/3D registration can often fail in practice: conventional optimization methods are prohibitively slow and susceptible to local minima, while neural networks trained on small datasets fail on new patients or require impractical landmark supervision. We present \name, a self-supervised approach that leverages patient-specific simulation and differentiable physics-based rendering to achieve accurate 2D/3D registration without relying on manually labeled data. Preoperatively, a CNN is trained to regress the pose of a randomly oriented synthetic \xray rendered from the preoperative CT. The CNN then initializes rapid intraoperative test-time optimization that uses the differentiable \xray renderer to refine the solution. Our work further proposes several geometrically principled methods for sampling camera poses from $\SE3$, for sparse differentiable rendering, and for driving registration in the tangent space~$\se3$ with geodesic and multiscale locality-sensitive losses. \name achieves sub-millimeter accuracy across surgical datasets at intraoperative speeds, improving upon existing unsupervised methods by an order of magnitude and even outperforming supervised baselines. Our implementation is at \url{https://github.com/eigenvivek/DiffPose}.
\end{abstract}

\section{Introduction}
Many high-stakes surgical procedures use intraoperative \xray image guidance to visualize surgical instruments and patient anatomy \cite{peters2001image}. While \xray imaging is both rapid and portable, it lacks the spatial detail of volumetric modalities such as CT. The advantages of 3D image guidance can be emulated by registering intraoperative \xrays to routinely acquired 3D preoperative scans, \ie estimating the intraoperative \xray scanner's pose relative to the patient's preoperative CT scan. Beyond localization of instruments relative to patient anatomy, accurate 2D/3D rigid registration is critical to cutting-edge surgical innovations incorporating robotic assistance and augmented reality \cite{adler1999image, andress2018fly, gao2020fiducial}.

Classical 2D/3D rigid registration methods apply ray tracing to CT volumes to render synthetic \xrays, referred to as digitally reconstructed radiographs (DRRs) in medical imaging \cite{aouadi2008accurate, van2011evaluation}. Iterative optimization is used to find the camera pose that generates a synthetic \xray that most closely matches the real \xray as quantified by an image similarity score. A major limitation of intensity-based methods is their limited capture range and sensitivity to the initial pose estimate \cite{unberath2021impact}. If the initial pose is even a few millimeters from the true pose, these methods can converge to a wrong solution. To this end, two methods for initial pose estimation are commonly deployed: landmark-based localization and CNN-based pose regression. 

In landmark-based localization, feature extractors find correspondences between anatomical landmarks in 2D and 3D images, which are then used by a Perspective-n-Point (PnP) solver to estimate the camera pose \cite{lepetit2009ep, li2012robust}. Such methods require expert knowledge of landmarks that are visible on \xray for the specific surgical application and manual annotation of 3D landmarks for every new CT volume \cite{grupp2020automatic}. While landmark localizers can be learned, only a few small labeled datasets exist for training deep networks \cite{grupp2019pose, pernus20133d, tomaevivc2002gold}, preventing the generalization of these methods to new patients, procedures, and pathologies.

In parallel, several approaches similar to PoseNet~\cite{kendall2015posenet} have trained CNNs to directly regress the pose of an intraoperative \xray \cite{miao2016cnn, bui2017x}. However, landmark-based localization consistently outperforms pose regression \cite{brachmann2018learning, walch2017image} due to CNNs performing pose regression via image retrieval instead of leveraging 3D structure \cite{sattler2019understanding}. Therefore, generalizing supervised pose regression across all patients would require unattainably large datasets. Instead, we train a \textit{patient-specific} pose regression CNN on potentially infinite synthetic \xrays rendered from the patient's own preoperative CT. We sample poses from a continuous distribution on $\SE3$ whose support includes all intraoperative views.

Even when supervised training data is augmented with synthetic \xrays, landmark-based localization and pose regression fail to achieve consistent sub-millimeter accuracy \cite{esteban2019towards, miao2016cnn, zhang2023patient}. Therefore, state-of-the-art 2D/3D registration methods refine initial pose estimates with intensity-based iterative optimization at test time (\ie intraoperatively). Unfortunately, this refinement can itself fail due to limitations with standard intensity losses (\eg MSE, SSIM, NCC). We improve this test-time optimization procedure in two ways: We use differentiable \xray rendering to optimize camera poses in the Lie algebra $\se3$, and we develop a sparse multiscale variant of local normalized cross correlation~\cite{avants2011reproducible, balakrishnan2018unsupervised} that is robust to local minima and faster to evaluate.

\subpara{Contributions.} We present \name, a self-supervised framework for differentiable 2D/3D registration (\Cref{fig:teaser}). First, we pretrain a patient-specific CNN with our self-supervised pose regression task, removing the need for manually annotated training data. Through a combination of image similarity and $\SE3$-geodesic losses, the CNN learns to predict accurate initial pose estimates. Then, using a differentiable \xray renderer for test-time optimization, we quickly refine the estimated camera pose by maximizing multiscale image similarity computed over sparse image patches. Our framework requires only a single preoperative CT scan, which is routinely acquired in the clinical standard of care. We evaluate our method on two publicly available datasets from different surgical specialties. While previous methods fail to generalize to new patients or procedures, \name consistently achieves sub-millimeter registration accuracy across populations and anatomical structures.

\begin{figure*}[t]
    \centering
    \includegraphics[width=\linewidth]{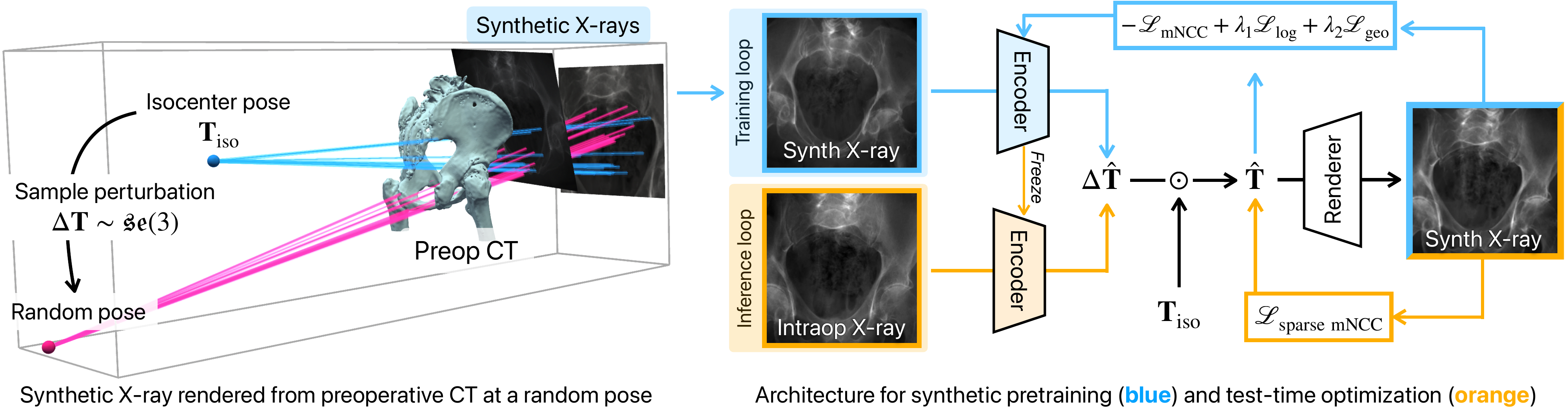}
    \caption{\textbf{\name setup.} \textit{Left:} Camera poses are sampled via random perturbations from the isocenter pose $\mathbf T_\mathrm{iso}$. \textit{Right:} An encoder is trained to regress the pose of a synthetic \xray using a combination of image similarity and $\SE3$-geodesic losses. At inference, the pose of a real intraoperative \xray is estimated by the encoder and iteratively refined using test-time optimization with differentiable rendering.}
    \label{fig:architecture}
\end{figure*}

\section{Related Work}
\noindent\textbf{Intensity-based 2D/3D registration.}
A 2D intraoperative \xray can be registered to a 3D preoperative CT volume by maximizing its similarity to a rendered \xray with respect to the camera pose \cite{maes1997multimodality, berger2016marker, chen2012automatic}. As traditional renderers are non-differentiable, pose estimation is commonly driven by gradient-free optimization~\cite{grupp2019pose}. This is both prohibitively slow for intraoperative use and often converges poorly in practice~\cite{zollei20012d}. On synthetic data, gradient-based optimization with DiffDRR, a differentiable renderer, has been shown to recover the true pose faster and more robustly than gradient-free methods~\cite{gopalakrishnan2022fast, abumoussa2023machine}. We use DiffDRR to develop a self-supervised method for differentiable registration and evaluate on real clinical \xray and CT images. 

\subpara{Landmark-based 2D/3D registration.}
Given supervision via anatomical landmarks, a camera pose can be estimated by applying a PnP solver~\cite{fischler1981random} to corresponding landmarks extracted from 2D intraoperative and 3D preoperative imaging. Recent approaches train neural networks to segment intraoperative \xrays and localize predefined landmarks from the segmentation masks \cite{grupp2020automatic, mizoe20214d, esteban2019towards, jaganathan2021deep, nguyen2022automatic}. This approach requires expert knowledge of which landmarks to extract and manual annotation of a large 2D/3D dataset for the anatomy of interest, precluding application to new anatomical structures. These approaches often fail on patients with poses, anatomies, or pathologies that are substantially different from those in the training set. 
% In contrast, our method does not use manual or estimated landmarks.
In contrast, our method does not use manual or estimated landmarks, operating directly on patient-specific image intensities.
% In contrast, our method does not use manual or estimated landmarks and operates directly on patient-specific image intensities.

\subpara{Camera pose regression.}
Pose estimation can also be formulated as a supervised regression problem \cite{miao2016cnn, bui2017x}. 
% Supervised regression models are trained with small annotated datasets of \xray images and associated camera poses from multiple patients. 
Prior work has focused on regressing parameterizations of rotations that have been shown to introduce discontinuities in the representation space, \eg Euler angles and quaternions~\cite{zhou2019continuity}. In point cloud registration, higher-dimensional parameterizations of $\SO3$ have been proposed that do not suffer from this issue \cite{zhou2019continuity,peretroukhin2020smooth,lin2023algebraically}. We investigate these parameterizations in the context of 2D/3D image registration.

\subpara{Self-supervision.}
Augmenting training data with synthetically generated \xrays helps overcome the small clinical sample sizes when training landmark extractors \cite{esteban2019towards, jaganathan2023self, shrestha2023x} and pose regressors \cite{miao2016cnn, zhang2023patient}. While generating synthetic data for PnP registration pretraining still requires manual annotations of landmarks on each new CT, rendered \xrays come with ground truth camera poses for free. Typically, a finite number of synthetic \xrays is generated for pretraining \cite{miao2016cnn, zhang2023patient}, failing to address the central limitation of CNN-based pose regression: as CNN-based pose regressors perform pose estimation through memorization of the training set~\cite{sattler2019understanding}, we use a fast differentiable \xray renderer to generate unlimited synthetic training data.

\section{Preliminaries}
Let $\mathbf V: \mathbb R^3 \to \mathbb R$ represent a 3D anatomical structure and $\mathbf I: \mathbb R^2 \to \mathbb R$ represent a 2D \xray of $\mathbf V$ taken at an unknown camera pose ${\mathbf T} \in \SE3$. These functions are related by the projection operator $\mathcal P(\mathbf T) : \mathbf V \to \mathbf I$, which models \xray image formation using a pinhole camera whose intrinsic parameters are known. Given $\mathbf I$ and $\mathbf V$, the goal of 2D/3D registration is to estimate the unknown pose $\mathbf T$.

\subpara{Differentiable rendering of synthetic \xrays.}
We render synthetic \xrays via a physics-based simulation of the image formation model. Specifically, we model the linear attenuation of \xrays in tissue and ignore second-order effects such as scattering and beam hardening \cite{staub2013digitally}. Let $\mathbf r(\alpha) = \mathbf s + \alpha (\mathbf p - \mathbf s)$ be a ray cast from the radiation source $\mathbf s \in \mathbb R^3$ to a pixel $\mathbf p \in \mathbb R^3$ on the imaging plane, where the spatial location of $\mathbf p$ relative to $\mathbf s$ is given by the known intrinsic parameters of the imaging system (\eg focal length).
As $\mathbf r$ travels through the anatomic volume $\mathbf V$, it loses intensity proportional to the linear attenuation coefficient $\mathbf V(\mathbf x)$ of every point $\mathbf x \in \mathbb R^3$ along its path. Assuming the ray has initial intensity $I_0$, its attenuated intensity once it has reached~$\mathbf p$ is given by the Beer-Lambert law~\cite{swinehart1962beer}:
\begin{align}
    \label{eq:image-formation-model}
    I(\mathbf p)
    &= I_0 \exp \Big( -\medint\int_{\mathbf x \in \mathbf r} \mathbf V(\mathbf x) \,\mathrm d \mathbf x \Big) \\
    &= I_0 \exp \Big( -\medint\int_0^1 \mathbf V\big(\mathbf r(\alpha)\big) \| \mathbf r'(\alpha)\| \,\mathrm d\alpha \Big) \\
    \label{eq:image-formation-model-attenuation}
    &= I_0 \exp \big( -I_\mu(\mathbf p) \big),
\end{align}
where $I_\mu(\mathbf p) \triangleq \| \mathbf p - \mathbf s \| \int_0^1 \mathbf V\big(\mathbf s + \alpha(\mathbf p - \mathbf s)\big) \,\mathrm d\alpha$ is proportional to the total energy absorbed by $\mathbf V$. We model the line integral in \cref{eq:image-formation-model-attenuation} by approximating $\mathbf V$ with a discrete 3D CT volume (\ie a voxel grid of linear attenuation coefficients). The discrete line integral is computed as in \cite{siddon1985fast}: 
\begin{equation}
    \label{eq:drr-image-formation}
    \| \mathbf p - \mathbf s \|
    \sum_{m=1}^{M-1} \mathbf V \left[ \mathbf r \left( \frac{\alpha_{m+1} + \alpha_m}{2} \right) \right] (\alpha_{m+1} - \alpha_m) ,
\end{equation}
where $\alpha_m$ parameterizes the locations where $\mathbf r$ intersects one of the orthogonal planes comprising the CT volume and $M$ is the number of such intersections. For a pixel grid $\mathbf P \in \mathbb R^{n \times 3}$ that forms the imaging plane, we can transform $\mathbf P$ and $\mathbf s$ by a rigid transformation $\mathbf T \in \SE3$ and render synthetic \xrays from any camera pose, \ie $\mathbf I = \mathcal P(\mathbf T) \circ \mathbf V$. The rendering equation \labelcref{eq:drr-image-formation} can be implemented as a series of vectorized tensor operations \cite{gopalakrishnan2022fast}, enabling the rendering of synthetic \xrays that are differentiable with respect to $\mathbf T$.

\subpara{Lie theory of $\SE3$.}
$\SE3$ is the Lie group of all rigid transformations in 3D. ${\mathbf T} \in \SE3$ comprises a rotation \mbox{$\mathbf R(\bm\varphi) \in \SO3$} and a translation $\mathbf t \in \mathbb R^3$, where $\bm\varphi \in \mathbb R^d$ is a Euclidean parameterization of $\SO3$. Classical parameterizations such as the axis-angle representation ($d=3$), Euler angles ($d=3$), and quaternions ($d=4$) suffer from degeneracies that can make pose estimation difficult, such as Gimbal lock or discontinuities in the representation space~\cite{zhou2019continuity}. This has motivated the development of several higher-dimensional ($d>5$) alternatives with theoretical advantages \cite{zhou2019continuity, lin2023algebraically, peretroukhin2020smooth}. Instead of parameterizing rotations and translations separately, we propose performing 2D/3D registration directly in the Lie algebra $\se3$, which is isomorphic to $\mathbb R^3 \times \mathbb R^3$ and jointly represents an axis-angle rotation about a translation vector \cite{murray2017mathematical}. We find that pose estimation and gradient-based optimization are most accurate when performed in $\se3$. Further, we leverage the Lie algebra to define geodesic losses for training pose regression CNNs and to parameterize random distributions over $\SE3$. \Cref{app:lie-theory} summarizes the relevant Lie theory.
% As such, new parameterizations have been developed for pose estimation tasks in machine learning (\eg Rotation6D \cite{zhou2019continuity}, Rotation10D \cite{peretroukhin2020smooth}, and the quaternion adjugate \cite{lin2023algebraically}). We compare the effectiveness of these different parameterizations in our experiments. 
% Our work considers parameterizing the pose estimation problem with the Lie algebras $\so3$ and $\se3$ corresponding to $\SO3$ and $\SE3$. We also leverage these Lie algebras to define geodesic distances between poses and parameterize distributions of random poses in $\SE3$.

\section{Methods}
Let $\mathcal I$ be the set of \xray images (both synthetic and real). Let $\mathcal E : \mathcal I \to \mathbb R^d \times \mathbb R^3$ be an encoder network that maps an \xray to a Euclidean parameterization of its camera pose. For every parameterization, there exists a surjective mapping $g: \mathbb R^d \times \mathbb R^3 \to \SE3$. \Eg, for $\se3$, $g(\cdot) = \exp(\cdot)$. Let $\mathcal P$ be a differentiable renderer that generates synthetic \xrays from a CT volume at any camera pose $\mathbf T \in \SE3$. We implement $\mathcal E$ as a convolutional neural network and $\mathcal P$ using DiffDRR~\cite{gopalakrishnan2022fast} such that the entire 2D/3D registration framework is end-to-end differentiable (\Cref{fig:architecture}). 

\subsection{Training Pose Estimation Networks}

\noindent\textbf{Sampling synthetic \xray poses.} The dimensions and voxel spacing of a given CT volume depend on the patient's anatomy, the type of surgery, and the imaging equipment available at the given hospital. Therefore, instead of sampling random camera poses in absolute coordinates, we sample random perturbations relative to the patient's isocenter pose $\mathbf T_{\mathrm{iso}}$. The isocenter is defined as the posteroanterior (PA) view where the camera is pointed directly at the patient (\ie $\mathbf R_{\mathrm{iso}} = \mathbf I$). The isocenter translation is $\mathbf t_{\mathrm{iso}} = (b_x \Delta_x, b_y \Delta_y, b_z \Delta_z) / 2$ where $(b_x, b_y, b_z)$ are the number of voxels in each dimension of the CT volume and $(\Delta_x, \Delta_y, \Delta_z)$ are the spacings of each voxel in millimeters per voxel. We sample the three rotational and three translational parameters of $\se3$ from normal distributions defined with sufficient variance to capture wide perturbations from the isocenter. Applying the exponential map to samples from the Lie algebra yields the perturbation $\Delta \mathbf T \in \SE3$. The random camera pose is then $\mathbf T = \Delta \mathbf T \circ \mathbf T_{\mathrm{iso}}$.

\subpara{Synthetic pose regression pretraining.}
After a preoperative CT scan has been obtained for a patient who will undergo surgery, we begin training $\mathcal E$. The training data consists solely of synthetic \xrays generated from the patient's preoperative CT at random camera poses. Given a randomly sampled camera pose $\mathbf T \in \SE3$, we render the associated synthetic \xray $\mathbf I = \mathcal P(\mathbf T) \circ \mathbf V$. From this image, we estimate the perturbation $\Delta \mathbf{\hat T} \triangleq \mathcal E(\mathbf I)$, construct the estimated pose $\mathbf{\hat T} = \Delta \mathbf{\hat T} \circ \mathbf T_{\mathrm{iso}}$, and render the predicted \xray $\mathbf{\hat I} = \mathcal P( \hat{\mathbf T}) \circ \mathbf V$. We optimize the weights of $\mathcal E$ using a combination of geodesic losses on $\SE3$ between $\mathbf{\hat T}$ and $\mathbf T$ and image-based losses on $\mathbf{\hat I}$ and $\mathbf I$.
% We update the weights of $\mathcal E$ with a combination of image-based loss functions and geodesic distances on $\SE3$.

\subsection{Registration Losses}
\label{subsec:losses}

\noindent\textbf{Geodesic pose regression losses.}
\label{sec:geodesics}
We use the geodesic distance between the estimated and ground truth camera poses as a regression loss when training $\mathcal E$. 
%We first describe the angular distance, an intuitive geodesic distance on $\SO3$, and then extend it to $\SE3$. 
Given two rotation matrices $\mathbf R_A, \mathbf R_B \in \SO3$, the angular distance between their axes of rotation is
\begin{align}
    d_\theta(\mathbf R_A, \mathbf R_B) 
    \label{eq:geodesic-so3}
    &= \arccos \left( \frac{\mathrm{trace}(\mathbf R_A^T \mathbf R_B) - 1}{2} \right) \\
    \label{eq:geodesic-log-so3}
    &= \| \log (\mathbf R_A^T \mathbf R_B) \| ,
\end{align}
where $\log(\cdot)$ is the logarithm map on $\SO3$ \cite{huynh2009metrics}.
Using the logarithm map on $\SE3$, this generalizes to a geodesic loss function on camera poses ${\mathbf T}_A, {\mathbf T}_B \in \SE3$:
\begin{equation}
    \label{eq:geodesic-log-se3}
    \mathcal L_{\mathrm{log}}({\mathbf T}_A, {\mathbf T}_B) = \| \log({\mathbf T}_A^{-1} {\mathbf T}_B) \| .
\end{equation}
We can also formulate a geodesic distance on $\SE3$ with units of length. Using the camera's focal length $f$, we convert the angular distance in \cref{eq:geodesic-so3} to an arc length:
\begin{equation}
    d_\theta(\mathbf R_A, \mathbf R_B; f) = \frac{f}{2} d_\theta(\mathbf R_A, \mathbf R_B) .
\end{equation}
When combined with the Euclidean distance on the translations $d_t(\mathbf t_A, \mathbf t_B) = \| \mathbf t_A - \mathbf t_B \|$, this yields the \textit{double geodesic} loss on $\SE3$ \cite{chirikjian2015partial}:
\medmuskip=2mu
\begin{equation}
    \label{eq:geodesic-se3}
    \mathcal L_{\mathrm{geo}}({\mathbf T}_A, {\mathbf T}_B; f) = \sqrt{d^2_\theta(\mathbf R_A, \mathbf R_B; f) + d^2_t(\mathbf t_A, \mathbf t_B)} .
\end{equation}
\medmuskip=4mu
%We train $\mathcal E$ using \cref{eq:geodesic-log-se3,eq:geodesic-se3}. In \Cref{sec:experiments}, we also include in the ablation studies the commonly used $L_1$ or $L_2$ on embeddings of rotations.
% In addition to the geodesic loss functions in \cref{eq:geodesic-log-se3,eq:geodesic-se3}, our experiments in \Cref{sec:experiments} we compare the commonly used loss of $L_1$ or $L_2$.
% In practice, we train using both \cref{eq:geodesic-log-se3,eq:geodesic-se3}.
% In \Cref{sec:experiments}, we compare the commonly used loss of $L_1$ or $L_2$ on embeddings of the rotations to the geodesic loss functions in \cref{eq:geodesic-log-se3,eq:geodesic-se3}.

\subpara{Multiscale NCC.}
\label{sec:multiscale-ncc}
% Image alignment tasks such as intensity-based registration or neural rendering rely on pixel-wise losses (\eg MSE or MAE) or image-based losses (\eg SSIM or LPIPS) to quantify the similarity between two images $\mathbf I_A$ and $\mathbf I_B$. 
Global normalized cross correlation (NCC) is a widely-used metric used to quantify the similarity between two images $\mathbf I_A$ and $\mathbf I_B$:
\begin{equation}
    \label{eq:ncc}
    \mathrm{NCC}(\mathbf I_A, \mathbf I_B) = \frac{1}{NM} \sum_{i=1}^N \sum_{j=1}^M \mathbf Z_A[i, j] \mathbf Z_B[i, j] ,
\end{equation}
where $\mathbf Z = (\mathbf I - \mu(\mathbf I)) / \sigma(\mathbf I)$ is an $N \times M$ image normalized by its pixel-wise mean and standard deviation.
Alternatively, \cref{eq:ncc} can be evaluated on overlapping patches of $\mathbf I_A$ and $\mathbf I_B$ to capture local correlations in small regions~\cite{avants2011reproducible}. When it successfully converges, we find the best registration accuracy achieved by local NCC is an order of magnitude better than global NCC. However, the metric is also more unstable and frequently guides the test-time optimizer to a poor quality solution. Instead, we find averaging NCC over multiple patch sizes, a generalization known as multiscale NCC ($\mathcal L_{\text{mNCC}}$), to be both more accurate and more numerically stable. This image similarity metric consistently achieves sub-millimeter registration accuracy while successfully converging more reliably than local NCC. \Cref{sec:experiments} and \Cref{app:image-similarity-metrics} provide evaluations and visualizations of multiple image losses for this application.

\subpara{Composite pretraining loss.} We train $\mathcal E$ using the following loss function ($\lambda_{1}$ and $\lambda_2$ are hyperparameters):
\begin{equation}
    \label{eq:pretraining-loss}
    -\mathcal L_{\text{mNCC}}(\mathbf I, \hat{\mathbf I}) + \lambda_1 \mathcal L_{\mathrm{log}}(\mathbf T, \mathbf{\hat T}) + \lambda_2 \mathcal L_{\mathrm{geo}}(\mathbf T, \mathbf{\hat T}) .
\end{equation} 

\subpara{Sparse differentiable rendering.}
% The main drawback of multiscale NCC is its computation time. We accelerate the calculation of multiscale NCC using sparse rendering, . In particular, we compute global NCC without additional rendering cost when computing local NCC on a small number of patches. Encoder activations are used to identify important image patches, which we make more likely to be sampled for sparse rendering. Visualizations and speed benchmarks for sparse multiscale NCC are provided in the \Cref{app:sparse-mncc}.
% The main drawback of multiscale NCC is its computation time. We accelerate the calculation of multiscale NCC using sparse rendering, closely related to prior work that used sparse image patches to estimate mutual information between images \cite{zollei20012d}. Given an image, we use the activations of the encoder $\mathcal E$ to identify important image regions. After sampling pixel locations from the activation map, we render a fixed number of patches with a predefined patch size. The key insight of our method is that, in addition to estimating local NCC with this small number of patches, we can also estimate global NCC without additional rendering cost. This is achieved by also evaluating \cref{eq:ncc} with the sparse subset of rendered pixels. Visualizations and speed benchmarks for sparse multiscale NCC are provided in the \Cref{app:sparse-mncc}.
Evaluating NCC across multiple scales is computationally expensive. We accelerate this calculation by estimating multiscale NCC with sparse rendering. Using the activations of the encoder $\mathcal E$, we identify important regions of the image. During test-time optimization, we use the activation map to sample random subsets of image patches to render. In addition to estimating local NCC with this small number of patches, we can also estimate global NCC without additional rendering cost. This is achieved by evaluating \cref{eq:ncc} with the sparse subset of rendered pixels. Visualizations and speed benchmarks for sparse multiscale NCC are provided in the \Cref{app:sparse-mncc}.

\subsection{Test-Time Optimization}
At test time (\ie during the surgery), the model is applied to real \xrays for which the ground truth pose $\mathbf T$ is unknown. Given a real \xray $\mathbf I$ acquired intraoperatively, the encoder estimates the underlying pose $\mathbf{\hat T}$. Despite training on a massive amount of synthetic data, this pose estimate often does not achieve optimal alignment on real images. %, further demonstrating the limitations of CNN-based approaches for camera pose regression. 
Further refining this initial pose estimate with intensity-based registration decreases registration error by orders of magnitude. To perform test-time optimization, we render the synthetic \xray $\mathbf{\hat I} = \mathcal P(\mathbf{\hat T}) \circ \mathbf V$ and compute an image-based similarity metric between the real and synthetic \xrays. Since $\mathcal P$ is differentiable, we can use gradient-based optimization to update $\mathbf{\hat T}$ such that the similarity between $\mathbf I$ and $\mathbf{\hat I}$ is increased. Note that gradient updates occur in the Euclidean representation space $\mathbb R^3 \times \mathbb R^3$ such that the estimated pose remains on the $\SE3$ manifold.

\begin{figure}[t]
    \centering
    \includegraphics[width=\linewidth]{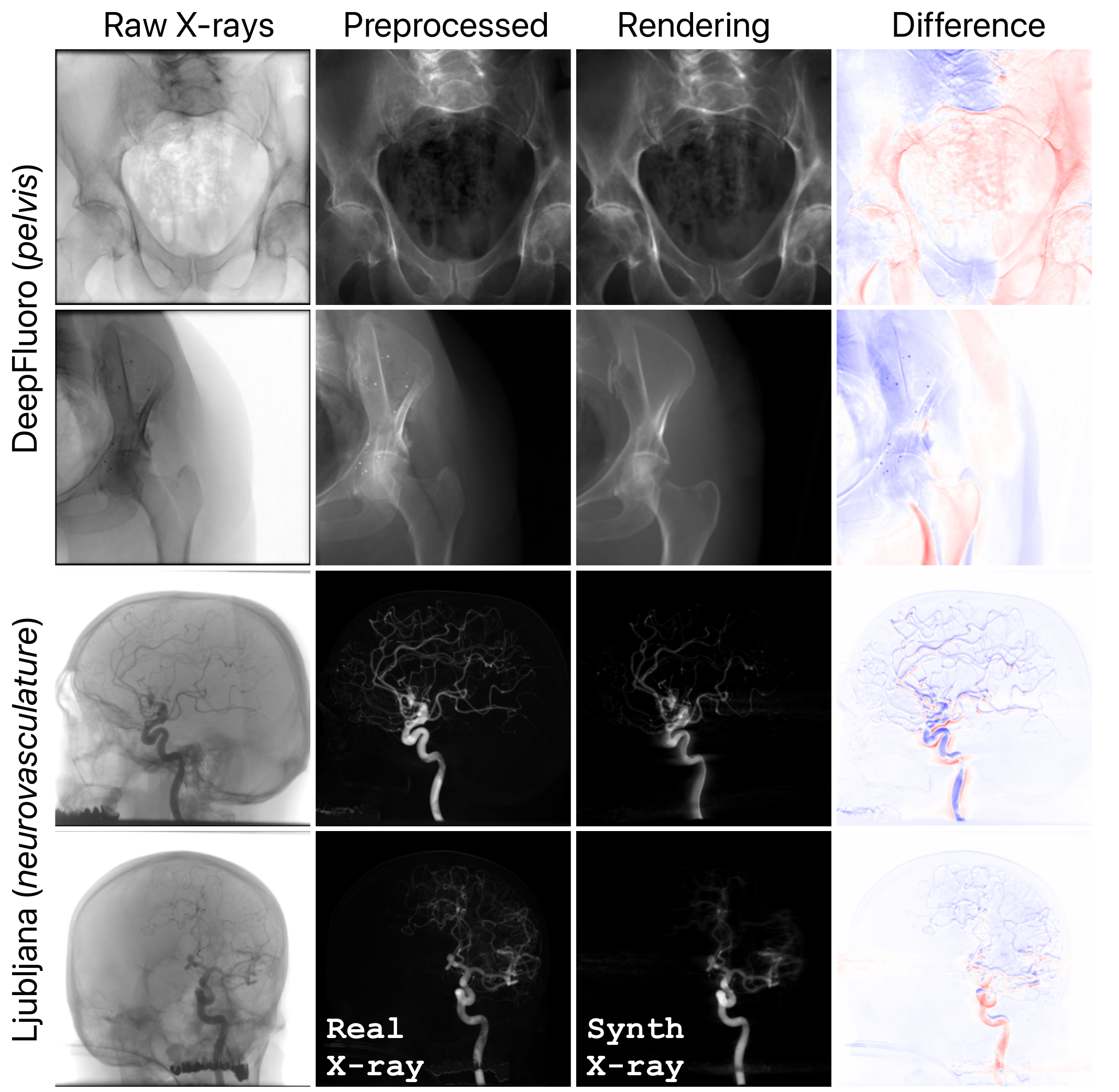}
    \caption{\textbf{Sample renders.} Raw \xrays are preprocessed to match the image formation model in~\cite{gopalakrishnan2022fast}. Difference maps between intraoperative \xrays and renderings from a preoperative CT visualize domain shift between real and synthetic images. In Row 2, the left femur moves between acquisition of preoperative and intraoperative images; in Rows 3 and 4, 3D volumes do not capture the smallest cranial blood vessels \cite{luo2014low}, so they cannot be rendered.}
    \label{fig:examples}
\end{figure}

\subsection{Landmark-Based Evaluation}
We evaluate performance with mean Target Registration Error (mTRE), an independent measurement of registration accuracy at important anatomical landmarks. For each CT volume, the open-source datasets we use in our experiments predefine a set of $m$ 3D anatomical landmarks $\mathbf M \in \mathbb R^{3 \times m}$. Since the intrinsic matrix $\mathbf K \in \mathbb R^{3 \times 3}$ of each imaging system used in these studies is known, we can calculate the perspective projection of $\mathbf M$ for any camera pose. mTRE is defined as the average distance between the projections given by the ground truth and estimated camera poses:
\begin{equation}
    \label{eq:mtre}
    \mathrm{mTRE}(\mathbf T, \mathbf{\hat T}) = \frac{1}{m} \| \mathbf K (
        [\mathbf R \mid \mathbf t] - [\mathbf{\hat R} \mid \mathbf{\hat t}]
    ) \mathbf M \|_F .
\end{equation}
Since we never use landmark supervision for training or test-time optimization, mTRE is an independent metric of registration accuracy. 
Following guidelines from the American Association of Physicists in Medicine (AAPM)~\cite{brock2017use}, a registration is successful if $\text{mTRE} \leq \SI{1}{\mm}$. We report a sub-millimeter success rate (\textbf{SMSR}) for all experiments.
% we define successful registration as sub-millimeter accurate and provide a sub-millimeter success rate (\textbf{SMSR}) in all experiments.

\subsection{Implementation Details}

\noindent\textbf{Pretraining.}
We used a ResNet18 backbone to implement the encoder $\mathcal E$. Extracted features were processed by two fully connected layers, which regressed the rotational ($\mathbb R^3$) and translational ($\mathbb R^3$) components of the pose, respectively. For each CT scan, a patient-specific $\mathcal E$ was trained from scratch using synthetic \xray images rendered on the fly. The Adam optimizer with warm-up was used with a maximal learning rate of $1 \times 10^{-3}$ and a cosine learning rate scheduler. As both the rendering of synthetic \xrays and CNN regression were performed on the same device, we were limited to batch sizes of eight $256 \times 256$ images on an A6000 GPU, leading us to replace batch normalization with group normalization~\cite{wu2018group}. $\mathcal E$ was trained on 1,000,000 synthetic images, which took approximately 12 hours. 
Lastly, we set $\lambda_1 = \lambda_2 = 10^{-2}$ in \cref{eq:pretraining-loss} and used multiscale NCC with patch sizes 13 and 256 (\ie the whole image).

\subpara{Test-time optimization.}
Test-time optimization was performed to refine the initial pose estimate. We used sparse multiscale NCC with 100 patches and a patch size of 13 as the image-based loss function and parameterized the space of poses with $\se3$. Gradient-based pose updates were performed using the Adam optimizer with a learning rate of $7.5 \times 10^{-3}$ on the rotational components of $\se3$ and a learning rate of $7.5 \times 10^0$ on the translational components of $\se3$ for 250 iterations. Additionally, a step learning rate decay at a factor of 0.9 was applied every 25 iterations. 

\section{Experiments}
\label{sec:experiments}

\begin{figure*}[t]
    \centering
    \includegraphics[width=\linewidth]{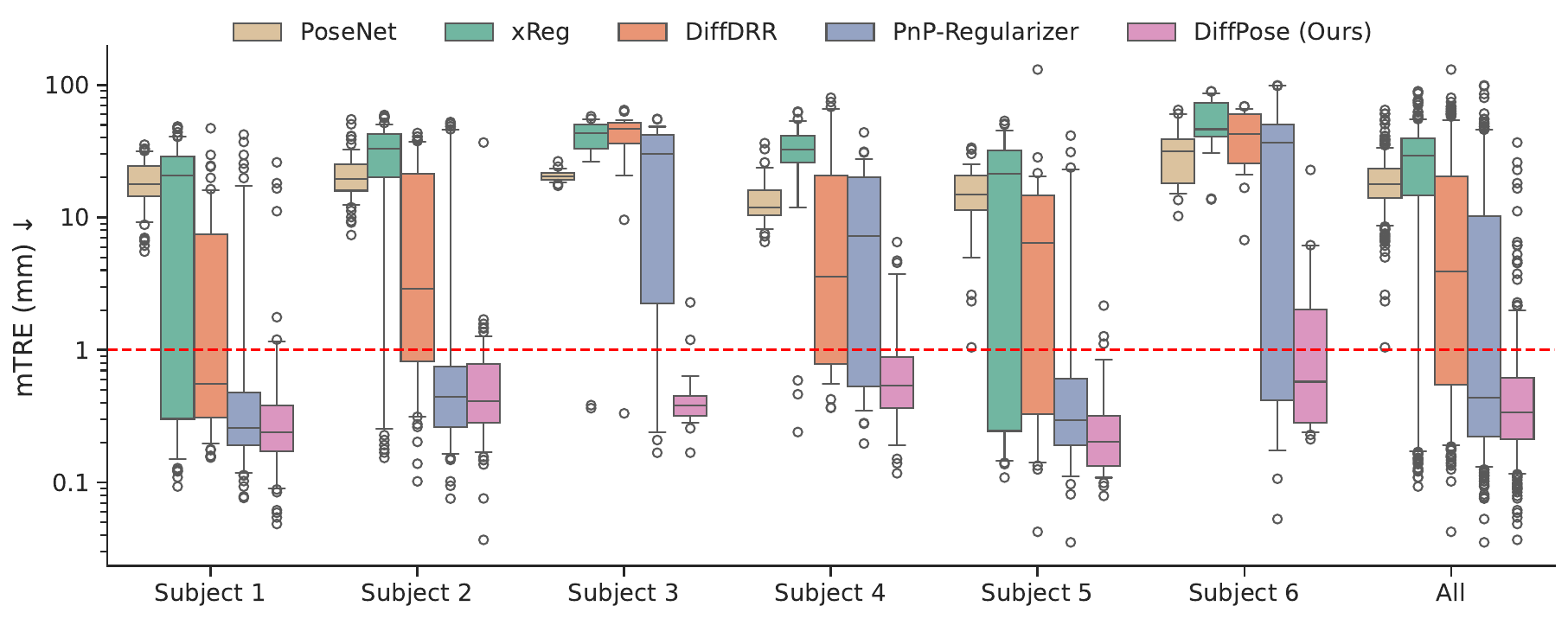}
    \caption{\textbf{Quantitative evaluation.} Evaluation of different registration methods on the DeepFluoro dataset via mTRE. A method successfully registered an \xray if the final mTRE was less than one millimeter ({\color{red}red line}). \name is the only method that consistently achieves sub-millimeter mTRE, outperforming fully supervised methods (PoseNet and PnP-Regularizer). Note that the y-axis is on a log-scale.}
    \label{fig:mtre}
\end{figure*}

\subsection{Datasets}
We perform an in-depth analysis of two public datasets that provide calibrated camera poses and expert annotations used for evaluation. We first evaluate our method using the DeepFluoro dataset, an open-source collection of pelvic \xrays and CTs from six cadavers \cite{grupp2020automatic}. For each subject, there is one CT scan and between 24-111 \xrays for a total of six CTs and 366 \xrays. For each \xray, the ground truth extrinsic matrix is provided. The \xray imaging system is also calibrated, \ie its intrinsic matrix is known. To demonstrate the difficulty in adapting existing registration methods to new surgical procedures, we also perform evaluations on the Ljubljana dataset~\cite{pernus20133d}, a clinical dataset consisting of 2D and 3D \xray angiography images from 10 patients undergoing neurovascular surgery. Each patient has two 2D \xrays and one 3D volume with accompanying ground truth extrinsic and intrinsic camera matrices. To demonstrate generality across surgical domains, we focus our ablations on the DeepFluoro dataset and do not change hyperparameters or modeling decisions for the Ljubljana dataset. Importantly, we note that sample sizes are generally low in real 2D/3D surgical datasets due to the difficulty of acquiring expert annotations with calibrated poses in surgical settings, limiting the utility of supervised methods.

Raw images acquired intraoperatively measure \xray attenuation. To match our physics-based differentiable renderer, we converted these to \xray absorption images by inverting \cref{eq:image-formation-model}, yielding $I_\mu(\mathbf p) = \log I_0 - \log I(\mathbf p)$. We estimate the rays' initial intensity as $I_0 = \max I(\mathbf r)$, where the max is taken over all images in the dataset (\ie the pixel with maximum intensity corresponds to an \xray that intersected no anatomy and therefore experienced zero attenuation). Additionally, we cropped the attenuation images by 50 pixels on each side to remove the effect of the collimator. Finally, log transforming yields the absorption image. Raw attenuation \xrays, preprocessed absorption \xrays, and synthetic \xrays rendered from a preoperative CT at the ground truth camera pose are shown in \Cref{fig:examples}.

\begin{figure*}[!htb]
    \centering
    \includegraphics[width=\linewidth]{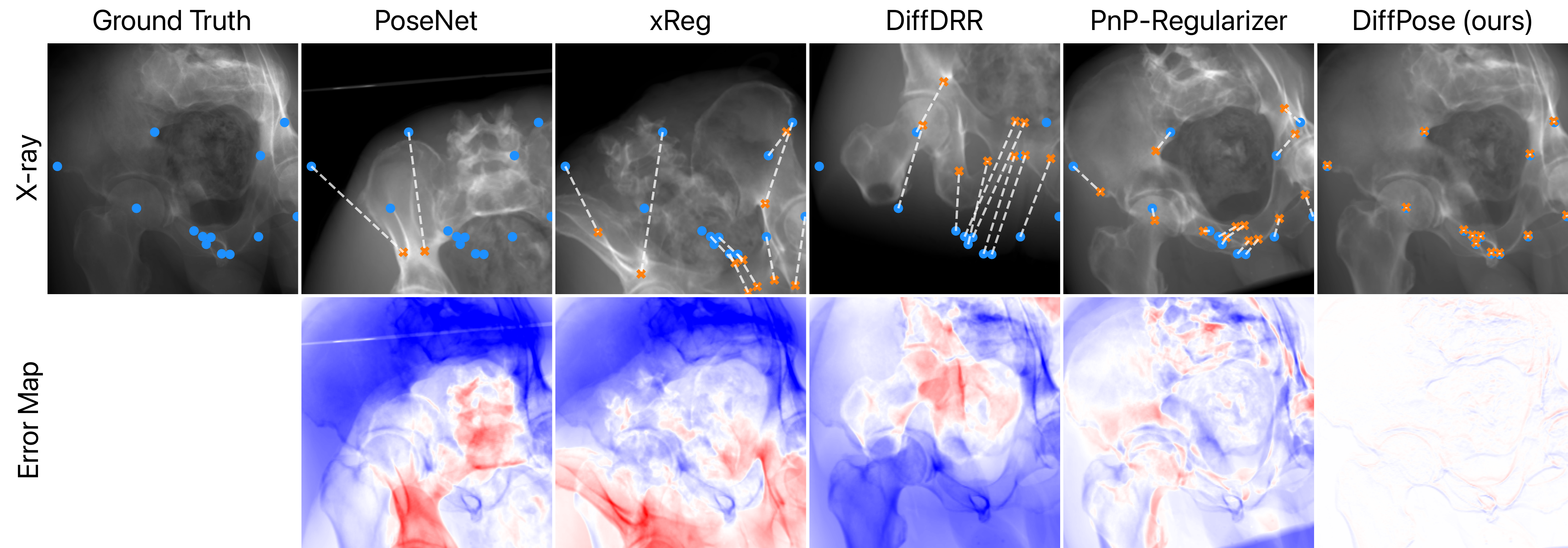}
    \caption{\textbf{Qualitative visualizations.} \textit{Top:} Renderings at the final pose estimates produced by different registration methods. Correspondences are drawn between true landmarks ({\color{cyan}blue}) and estimated landmarks ({\color{orange}orange}). \textit{Bottom:} To compare geometric alignment and not appearance, error maps are computed as the difference between the \xrays rendered at the ground truth pose and the final pose estimate.}
    \label{fig:qual}
\end{figure*}

Unlike many datasets used to evaluate previous 2D/3D registration methods, the CTs in the DeepFluoro and Ljubljana datasets are not reconstructed from the acquired \xrays. This makes our registration problem more challenging because the subject can move between the acquisition of the preoperative and intraoperative images, meaning the 2D \xrays are no longer directly embedded in the 3D CT scan. Testing on datasets with independent and physically-acquired 2D and 3D modalities avoids test set leakage during evaluation and directly simulates the clinical use case.

\subsection{Baseline Methods}
An evaluation of multiple existing 2D/3D registration approaches was conducted. We first compare the proposed method against PoseNet~\cite{kendall2015posenet}, a supervised pose regression algorithm. We evaluate PoseNet using leave-one-subject-out cross-validation, training a new model for each subset of five subjects. We also include two unsupervised intensity-based registration methods: xReg~\cite{grupp2019pose} and DiffDRR~\cite{gopalakrishnan2022fast}. Initialized at a PA pose, xReg uses a multi-stage gradient-free optimization routine and a patch-based image-gradient NCC similarity metric \cite{grupp2018patch}. DiffDRR was initialized at the same PA pose as xReg. We evaluated DiffDRR using the same learning rates and image-based loss function as our test-time optimization. Finally, we compare against two PnP registration methods \cite{grupp2020automatic}. In the first approach (PnP), a U-Net is trained to extract landmarks from 2D \xrays, again using a leave-one-subject-out approach. These landmarks are used with a PnP solver to predict the camera pose, which can be subject to poor initialization especially if less than four landmarks are visible in the 2D image \cite{grupp2020automatic}. A more robust approach, PnP-Regularizer, uses the detected landmarks to formulate a regularizer on potential camera poses, effectively optimizing estimated mTRE~\labelcref{eq:mtre}.

\subsection{Results}

\noindent\textbf{DeepFluoro dataset.} \Cref{fig:mtre} reports the mTRE statistics for each subject and in aggregate for DeepFluoro. Sub-millimeter success rates across baselines are summarized in \Cref{tab:success-rate}. Our encoder's estimates were within \SI{10}{\mm} of the true pose for 80\% of test cases, a commonly used cutoff for successful registration \cite{gao2023fully}, but only achieved \textit{sub-}millimeter mTRE on 1\% of test cases (\Cref{tab:ablations}). Refining the initial pose with test-time optimization was thus critical to achieving successful registration, leading to \successrate of test cases achieving sub-millimeter mTRE across all subjects.

\begin{table}[b]
\centering
\setlength{\tabcolsep}{5pt}
\caption{\textbf{Baseline comparisons.} 
Sub-millimeter success rate (SMSR) and runtime statistics for DeepFluoro subjects. Methods are classified by registration strategy: unsupervised intensity-based (I), supervised landmark-based (L), and pose regression (R).}
\label{tab:success-rate}
\begin{tabular}{@{}lccccc@{}}
\toprule
 & I & L & R & SMSR $\uparrow$ & Time (s) $\downarrow$ \\ \midrule
PoseNet \cite{kendall2015posenet} &  &  & \checkmark & {\color{white}0}0\% & N/A \\
xReg \cite{grupp2019pose} & \checkmark &  &  & 20\% & \SI{1.3 \pm 1.9}{} \\
DiffDRR (mNCC) \cite{gopalakrishnan2022fast} & \checkmark &  &  & 37\% & \SI{7.5 \pm 5.9}{} \\
PnP \cite{grupp2020automatic} &  & \checkmark &  & {\color{white}0}1\% & N/A \\
PnP-Regularizer \cite{grupp2020automatic} & \checkmark & \checkmark &  & 69\% & \textbf{\SI{1.1 \pm 1.6}{}} \\
\name (Ours) & \checkmark &  & \checkmark & \textbf{\successrate} & \SI{2.2 \pm 1.2}{} \\ \bottomrule
\end{tabular}
\end{table}

\begin{figure}[!hb]
    \centering
    \includegraphics[trim=0 0 0 2cm,width=0.975\linewidth]{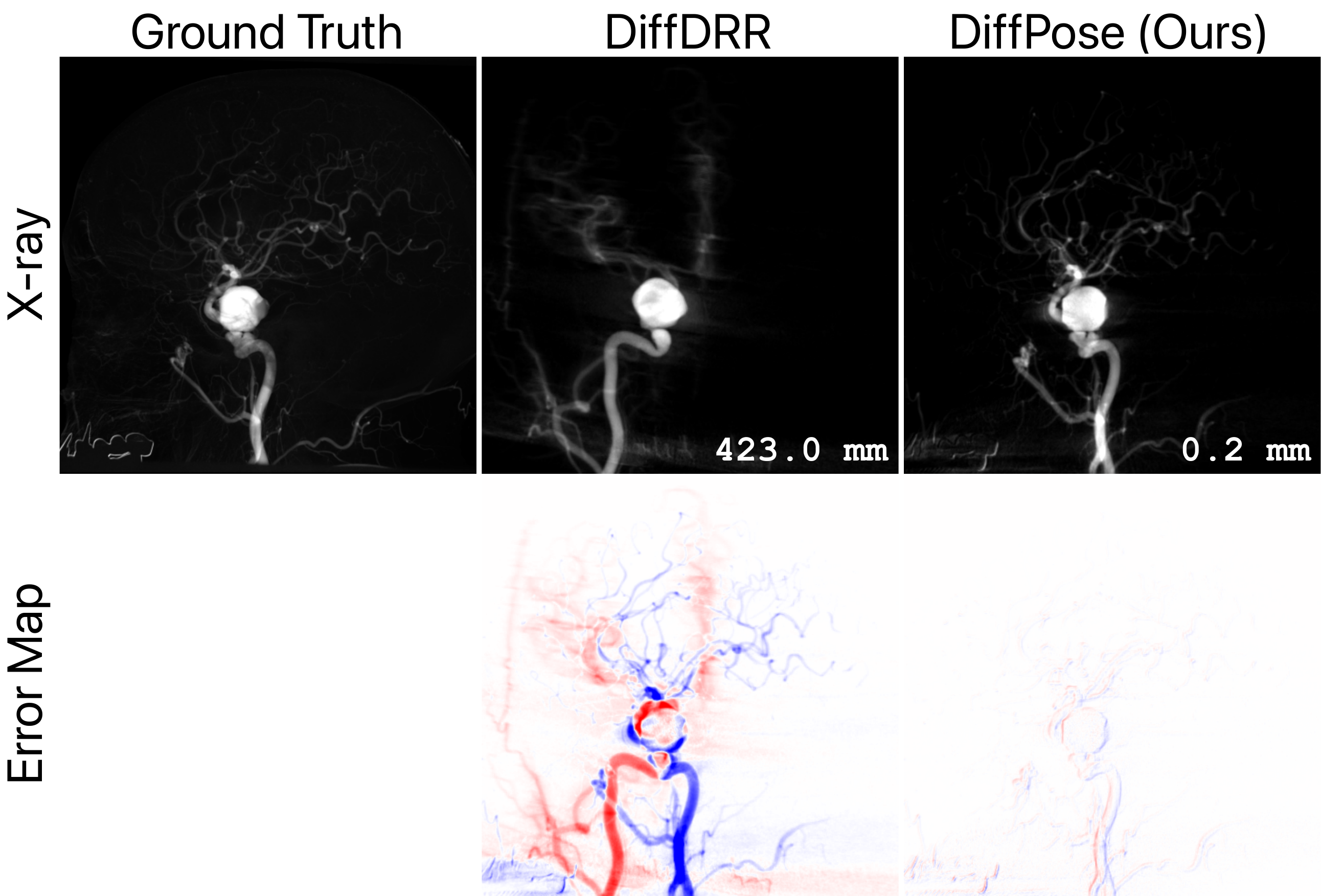}
    \caption{\textbf{External dataset validation.} Using modeling decisions from a pelvic dataset, \name demonstrates high registration accuracy on blood vessels in the brain. \textit{Top:} Renders at final pose estimates with associated mTRE. \textit{Bottom:} Error maps between \xrays rendered at the estimated and ground truth poses.}
    \label{fig:ljubljana}
\end{figure}

As subjects are not registered to a common reference frame and there is a limited number of \xrays, a purely supervised pose regressor (PoseNet~\cite{kendall2015posenet}) incurs large error. Traditional gradient-free optimization (xReg~\cite{grupp2019pose}) is able to successfully recover the poses of some \xrays but frequently converges to suboptimal minima. The gradient-based approach (DiffDRR~\cite{gopalakrishnan2022fast}) is more robust than gradient-free methods, but not to a surgically viable degree. Taken together, xReg and DiffDRR show that test-time optimization alone is insufficient for accurate intraoperative 2D/3D registration. PnP methods provide better initial pose estimates on average, but their accuracy diminishes when only a few landmarks are present in the 2D image. When combined with xReg, PnP-Regularizer successfully registered 69\% of test \xrays with high inter-subject variability: on Subjects 3, 4, and 6, \name consistently achieves sub-millimeter accuracy while PnP-Regularizer's median mTRE is around \SI{10}{\mm}. Inter-subject variability arises from out-of-distribution failures when the pose or appearance of a particular \xray does not match the training set, demonstrating the need for patient-specific methods. Visual comparisons are shown in \Cref{fig:qual}.

\subpara{Ljubljana dataset.}
Certain baselines are not applicable to the Ljubljana dataset. PnP-based methods cannot be trained as there are no segmentation masks with which to estimate landmarks, and the small sample size (20 \xrays) precludes training an accurate PoseNet. However, as our method does not require any labeled training data, we train patient-specific pose regression networks. We compared to DiffDRR~\cite{gopalakrishnan2022fast} initialized at a PA pose and found that \name achieved a median mTRE of \SI{0.2 \pm 10.6}{\mm} (85\% SMSR) as compared to DiffDRR's \SI{263.7 \pm 193.9}{\mm} (25\% SMSR). Qualitative results are visualized in \Cref{fig:ljubljana}. 
The high performance of \name on the neurovasculature, a highly distinct surgical domain, demonstrates its generalizability across patients and anatomical structures.

\subsection{Ablation Studies}

\noindent\textbf{Choice of pretraining loss.}
\Cref{tab:ablations} compares our composite loss \labelcref{eq:pretraining-loss} to isolated image (Row 3) and geodesic losses (Row 4), as well as an L2 loss on Euler angles (Row 5). All models were trained with the same number of synthetic images. We find that combining image and geodesic losses produces the most accurate model (Row 2).
\begin{table}[!t]
\centering
\setlength{\tabcolsep}{4pt}
\caption{Test time-optimization and pose regression loss ablations.}
\label{tab:ablations}
\begin{tabular}{@{}lccccc@{}}
\toprule
 & SMSR $\uparrow$ & mTRE (\si{\mm}) $\downarrow$ \\ \midrule
\name (Ours) & \textbf{\successrate} & \textbf{\SI{0.9 \pm 2.8}{}} \\ 
\midrule
Remove test-time opt. (TTOpt) & 1\% & \SI{5.4 \pm 4.3}{} \\
Remove TTOpt and $\mathcal L_{\mathrm{mNCC}}$ & 0\% & \SI{8.3 \pm 4.7}{} \\
Remove TTOpt, $\mathcal L_{\mathrm{geo}}$, and $\mathcal L_{\mathrm{\log}}$ & 0\% & \SI{18.1 \pm 4.9}{} \\
Remove TTOpt and use $L_2$ & 0\% & \SI{35.5 \pm 10.4}{} \\
\bottomrule
\end{tabular}
\vspace{-1em}
\end{table}

\subpara{Choice of $\SE3$ parameterization.}
\Cref{tab:parameterizations} compares optimizing camera poses directly in $\se3$ to more conventional parameterizations that treat rotational and translational components as independent. Despite previously reported deficiencies of Euler angles and quaternions in other contexts \cite{zhou2019continuity, peretroukhin2020smooth}, they are the most performant parameterizations after the proposed $\se3$ for 2D/3D registration.

\begin{table}[!b]
\vspace{-1em}
\centering
\setlength{\tabcolsep}{4pt}
\caption{Comparison of different parameterizations of $\SE3$ for test-time optimization with identical initialization.}
\label{tab:parameterizations}
\begin{tabular}{@{}lccc@{}}
\toprule
\textbf{} & SMSR $\uparrow$ & mTRE (\si{\mm}) $\downarrow$ \\ \midrule
$\se3$ (Ours)  & \textbf{\successrate} & \textbf{\SI{0.9 \pm 2.8}{}} \\
\midrule
Axis-angle $\times~\mathbb R^3$ & 75\% & \SI{1.1 \pm 4.6}{} \\
Euler angles $\times~\mathbb R^3$ & 83\% & \SI{1.0 \pm 3.2}{} \\
Quaternion $\times~\mathbb R^3$ & 83\% & \SI{1.0 \pm 3.7}{} \\
Rotation6D \cite{zhou2019continuity} $\times~\mathbb R^3$ &83\% & \SI{1.0 \pm 3.2}{} \\
Rotation10D \cite{peretroukhin2020smooth} $\times~\mathbb R^3$ & 81\% & \SI{1.5 \pm 4.8}{} \\
Quaternion Adjugate \cite{lin2023algebraically} $\times~\mathbb R^3$ & 77\% & \SI{3.5 \pm 9.7}{} \\ \bottomrule
\end{tabular}
\end{table}

\subpara{Choice of image similarity metric.}
\Cref{tab:metrics} compares sparse multiscale NCC (mNCC) to previously described loss functions for 2D/3D rigid registration. Local NCC~\cite{avants2011reproducible} performs competitively but has high variance, demonstrating the instability of the loss. Using mNCC, which averages global and local NCC, stabilizes the variance and results in the highest success rate (see \Cref{app:image-similarity-metrics,app:sparse-mncc}).

\begin{table}[!t]
\centering
\caption{Image similarity loss comparisons. All rows correspond to test time optimization with identical pretrained initialization.}
\label{tab:metrics}
\begin{tabular}{@{}lccc@{}}
\toprule
\textbf{} & SMSR $\uparrow$ & mTRE (\si{\mm}) $\downarrow$ \\ \midrule
Sparse mNCC (Ours)  & \textbf{\successrate} & \textbf{\bestmtre} &  \\
\midrule
Global NCC & 27\% & \SI{4.4 \pm 5.2}{} \\
Local NCC \cite{avants2011reproducible} & 81\% & \SI{1.7 \pm 5.2}{} \\
Gradient NCC \cite{grupp2018patch} & {\color{white}0}9\% & \SI{13.3 \pm 5.2}{} \\
SSIM & {\color{white}0}1\% & \SI{13.4 \pm 7.8}{} \\
MSE & {\color{white}0}0\% & \SI{30.8 \pm 15.8}{} \\
MAE & {\color{white}0}0\% & \SI{28.4 \pm 14.1}{} \\ \bottomrule
\end{tabular}
\end{table}

\section{Discussion}
\subpara{Limitations and future work.} DiffPose estimates a single rigid transformation between pre- and intraoperative scans, precluding direct application to deformable target structures. Fortunately, our core contributions can be extended to estimate piecewise rigid transformations (\eg one transform per rigid anatomical component), which can be aggregated to model arbitrary deformations~\cite{arsigny2009fast}. 
% Second, our one-ray-per-pixel ray tracing formulation yields aliasing artifacts in low-resolution \xrays, hindering faster registration at lower resolutions. These artifacts can be over
% Fortunately, our core contributions are straightforward to extend to arbitrary deformation models, enabling diverse surgical applications. Further, our one-ray-per-pixel ray tracing formulation yields aliasing artifacts on low-resolution \xrays that may hinder integration with coarse-to-fine registration frameworks. These artifacts can be overcome with multi-ray-per-pixel ray tracing at the cost of efficiency, or through the development of new ray tracing techniques that better model the thickness of rays cast to an individual pixel. 
Further, DiffPose uses per-subject training (analogous to NeRF~\cite{mildenhall2020nerf}), which may be too slow for emergency surgeries that do not allow for hours-long pretraining. We expect that our pose regression networks can be well-initialized by first training on a dataset of preregistered CT scans and synthetic \xrays, and then rapidly fine-tuning on a new subject in a few iterations of our patient-specific pretraining task (\eg with MAML~\cite{finn2017model}).

\subpara{Conclusion.} 
Intraoperative 2D/3D registration holds immense promise that has as yet been unfulfilled due to a high rate of registration failure and dependence on expert supervision. We present \name, the first intraoperative 2D/3D registration framework that is sub-millimeter accurate without using any surgically impractical supervision. \name demonstrated strong performance relative to current methods while retaining surgically relevant runtime, thus enabling successful applications to intraoperative pelvic and neurovascular procedures and many more surgical domains.

% \section*{Acknowledgments}
\subpara{Acknowledgements.} 
The authors thank Andrew Abumoussa, Nalini Singh, and Bill Worstell for helpful feedback. This work was supported, in part, by NIH NIBIB NAC P41EB015902, NIH NINDS U19NS115388, NIH NIBIB 5T32EB001680-19, and a Takeda Fellowship.

{\small \bibliographystyle{ieeenat_fullname} \bibliography{main}}

\maketitlesupplementary
\appendix

\section{Lie Theory of SE(3)}
\label{app:lie-theory}

We present a brief overview of the Lie theory of $\SE3$ as it pertains to our method. A camera pose $\mathbf T \in \SE3$ can be represented as the matrix
\begin{equation}
    \mathbf T =
    \begin{bmatrix}
        \mathbf R & \mathbf t \\
        \mathbf 0 & 1
    \end{bmatrix} \in \SE3 \,,
\end{equation}
where $\mathbf R \in \SO3$ is a $3 \times 3$ rotation matrix and $\mathbf t \in \mathbb R^3$ is a translation. The Lie group $\SE3$ corresponds to the Lie algebra $\se3$, spanned by six basis vectors representing either infinitesimal rotations or translations along a specific axis~\cite{blanco2021tutorial}:
\begin{equation}
    \begin{gathered}
        \mathbf G_1 = \begin{bmatrix}
            0 & 0 & 0 & 0 \\
            0 & 0 & -1 & 0 \\
            0 & 1 & 0 & 0 \\
            0 & 0 & 0 & 0 \
        \end{bmatrix} \quad
        \mathbf G_2 = \begin{bmatrix}
            0 & 0 & 1 & 0 \\
            0 & 0 & 0 & 0 \\
            -1 & 0 & 0 & 0 \\
            0 & 0 & 0 & 0 \
        \end{bmatrix} \quad
        \mathbf G_3 = \begin{bmatrix}
            0 & -1 & 0 & 0 \\
            1 & 0 & 0 & 0 \\
            0 & 0 & 0 & 0 \\
            0 & 0 & 0 & 0 \
        \end{bmatrix} \\
        \mathbf G_4 = \begin{bmatrix}
            0 & 0 & 0 & 1 \\
            0 & 0 & 0 & 0 \\
            0 & 0 & 0 & 0 \\
            0 & 0 & 0 & 0 \
        \end{bmatrix} \quad
        \mathbf G_5 = \begin{bmatrix}
            0 & 0 & 0 & 0 \\
            0 & 0 & 0 & 1 \\
            0 & 0 & 0 & 0 \\
            0 & 0 & 0 & 0 \
        \end{bmatrix} \quad
        \mathbf G_6 = \begin{bmatrix}
            0 & 0 & 0 & 0 \\
            0 & 0 & 0 & 0 \\
            0 & 0 & 0 & 1 \\
            0 & 0 & 0 & 0 \
        \end{bmatrix}
    \end{gathered}
\end{equation}
That is, any transformation $\mathbf T \in \SE3$ can be represented as a 6-vector $\mathbf v = \begin{bmatrix} \bm \omega & \mathbf u \end{bmatrix}^T$ corresponding to the matrix
\begin{equation}
    \label{eq:se3-log}
    \log \mathbf T 
    = \omega_1 \mathbf G_1 + \omega_2 \mathbf G_2 + \omega_3 \mathbf G_3 + u_1 \mathbf G_4 + u_2 \mathbf G_5 + u_3 \mathbf G_6
    = \begin{bmatrix}
        0 & -\omega_3 & \omega_2 & u_1 \\
        \omega_3 & 0 & -\omega_1 & u_2 \\
        -\omega_2 & \omega_1 & 0 & u_3 \\
        0 & 0 & 0 & 0
    \end{bmatrix} \in \se3 \,,
\end{equation}
where $\log: \SE3 \to \se3$ is the logarithmic map. In the literature, it is common to ``vee'' operator $(\cdot)^\vee$ to relate the matrix $\log \mathbf T$ to its vector representation $\mathbf v$. While most authors will write
\begin{equation}
    (\log \mathbf T)^\vee 
    = \begin{bmatrix}
        0 & -\omega_3 & \omega_2 & u_1 \\
        \omega_3 & 0 & -\omega_1 & u_2 \\
        -\omega_2 & \omega_1 & 0 & u_3 \\
        0 & 0 & 0 & 0
    \end{bmatrix}^\vee
    = \begin{bmatrix}
        \omega_1 & \omega_2 & \omega_3 & u_1 & u_2 & u_3
    \end{bmatrix}^T
    = \mathbf v \,,
\end{equation}
we let $\log(\cdot)$ implicitly denote this vectorization, in a slight abuse of notation. The matrix exponential provides the map $\exp : \se3 \to \SE3$. While it is common to use the analogous ``hat'' operator to represent $\mathbf T = \exp(\mathbf v^\wedge)$, we also treat this as implied. By grouping even and odd powers in their Taylor expansions, the logarithmic and exponential maps can be calculated in closed form~\cite{eade2013lie}. The implementations we use are available in PyTorch3D~\cite{ravi2020pytorch3d}.

An important interpretation appears when observing the equation for the $\exp$ map:
\begin{equation*}
    \mathbf T
    = \exp(\mathbf v^\wedge)
    = \begin{bmatrix}
        \exp \bm \omega^\wedge & \bm\Omega \mathbf u \\
        0 & 1
    \end{bmatrix}
    \quad\text{where}\quad
    \bm\Omega
    = \mathbf I + \left(\frac{1 - \cos\theta}{\theta^2}\right)\bm\omega^\wedge + \left(\frac{\theta - \sin\theta}{\theta^3}\right)(\bm\omega^\wedge)^2
\end{equation*}
and $\theta$ is given by \cref{eq:geodesic-log-so3}. Note, the translational component in $\mathbf T$ is produced through a combination of $\bm\omega$ and $\mathbf u$. That is, unlike all other Euclidean parameterizations of $\SE3$ considered in \Cref{tab:parameterizations}, the translational and rotational components are \textit{not} independent when represented in $\se3$. It is possible that the improved performance of $\se3$ is due to the coupling of rotational and translational components in the representation. Finally, note that the logarithmic map on $\SO3$, \ie $\log(\mathbf R) = \bm\omega$, is equivalent to the axis-angle representation \cite{eade2013lie}. Therefore, the parameterization in the second row of \Cref{tab:parameterizations} is the same as $\so3 \times \mathbb R^3$.

\clearpage
\section{Visualizing Loss Landscapes for Multiple Image Similarity Metrics}
\label{app:image-similarity-metrics}

\begin{figure*}[!ht]
    \centering
    \includegraphics[width=0.625\linewidth]{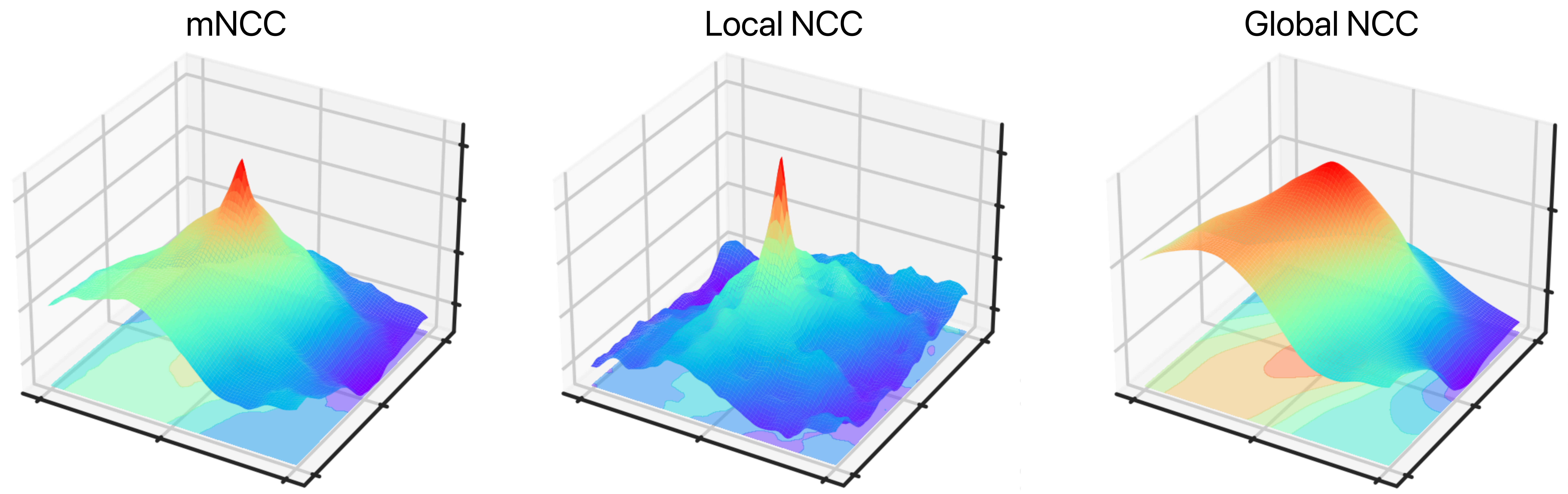}
    \caption{\textbf{Visual comparison of mNCC, local NCC, and global NCC.}}
    \label{fig:loss-landscapes-zoom}
\end{figure*}

In \Cref{fig:loss-landscapes-zoom}, we compare the loss landscapes of multiscale NCC (mNCC), local NCC (patch size of 13), and global NCC (\ie the whole image is a single patch). Loss landscapes are generated by measuring the similarity between the ground truth X-ray and synthetic X-rays rendered at perturbations from the ground truth camera pose. Local NCC has a sharp peak at the ground truth camera pose, however, its landscape has many local minima. In contrast, global NCC is much smoother, but has a less defined peak. Averaging local and global NCC yields mNCC, which has both a strong peak and a smooth landscape.

\begin{figure*}[!hb]
    \centering
    \includegraphics[width=0.975\linewidth]{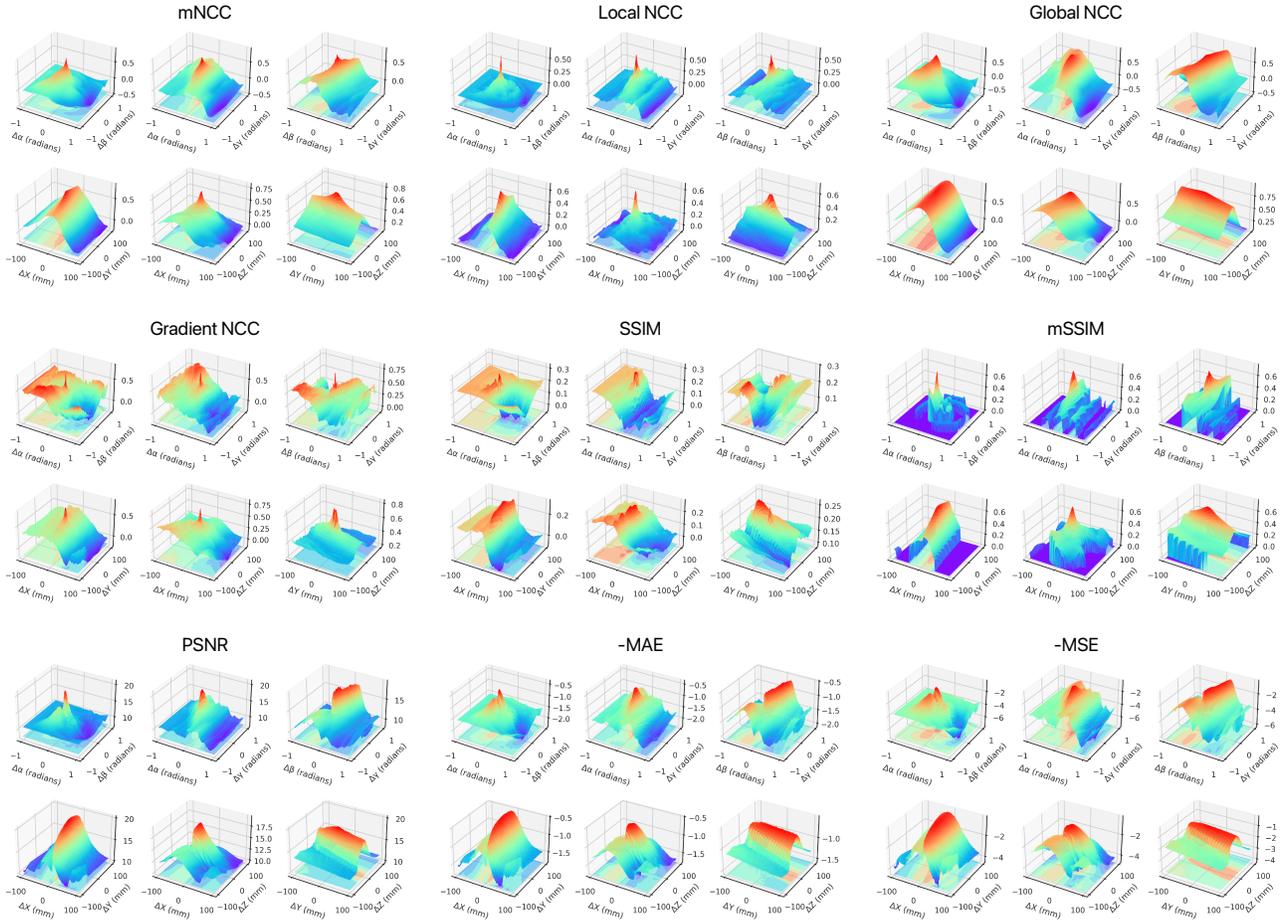}
    \caption{\textbf{Visualization of image-based loss landscapes.} mNCC is the most amenable to gradient-based optimization.}
    \label{fig:loss-landscapes}
\end{figure*}

In \Cref{fig:loss-landscapes}, we compare mNCC to the following image similarity metrics: local NCC, global NCC, gradient NCC \cite{grupp2018patch}, SSIM, multiscale SSIM (mSSIM), PSNR, negative MAE, and negative MSE. For the 6 \dof in a camera pose, rotational perturbations are jointly sampled from \SI{\pm 1}{\radian} and translational perturbations are jointly sampled from \SI{\pm 100}{\mm}.

\section{Visualizing Sparse Multiscale NCC}
\label{app:sparse-mncc}
\begin{figure}[!h]
    \centering
    \includegraphics[width=0.85\linewidth]{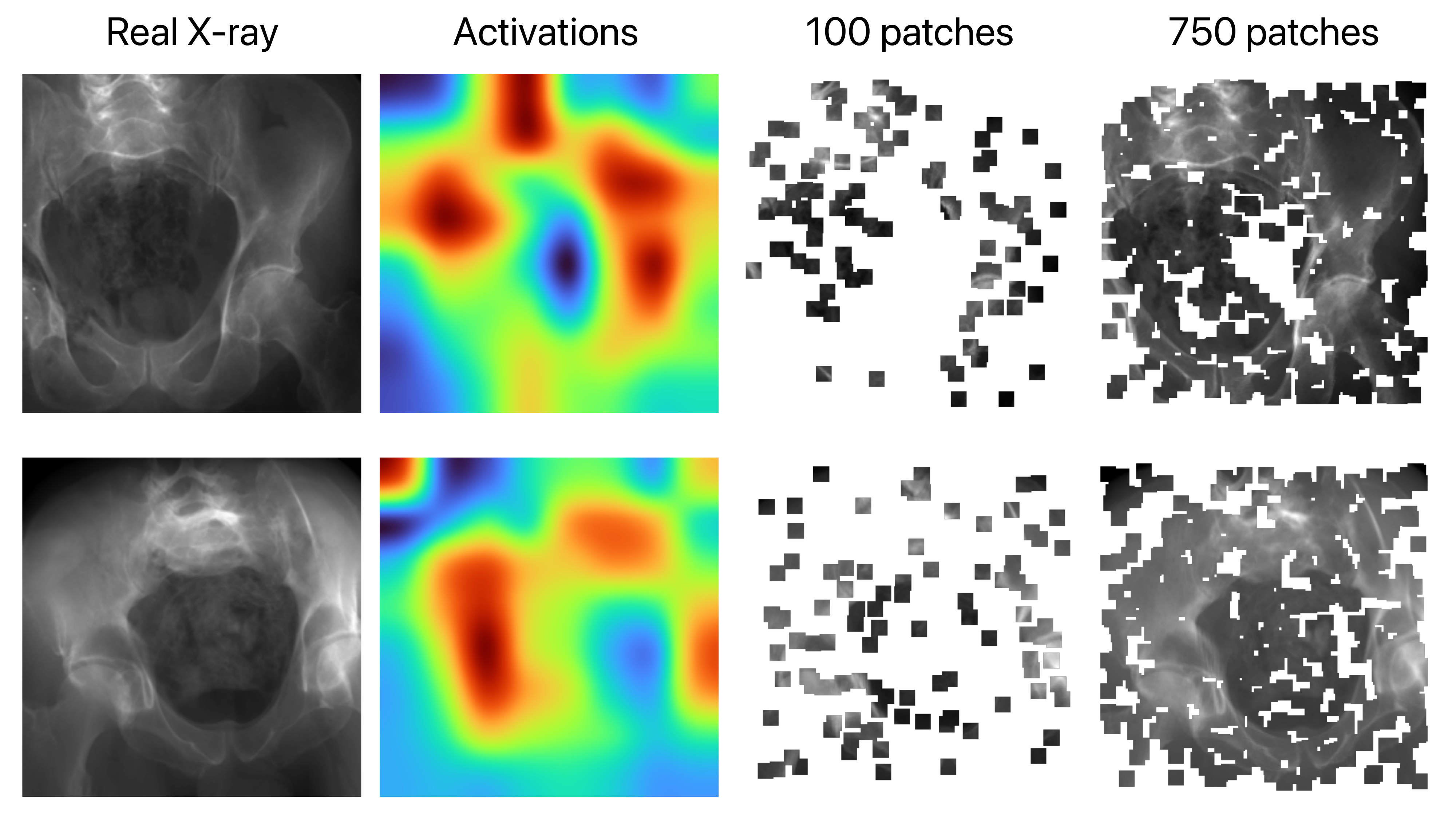}
    \caption{\textbf{Visualization of sparse mNCC.} We compute mNCC using sparse image patches rendered around anatomical structures that drive 2D/3D registration. In our experiments, we compute sparse mNCC using 100 patches with a patch size of 13.}
    \label{fig:sparse-mncc}
\end{figure}

\Cref{fig:sparse-mncc} visualizes the sparse patch-wise differentiable rendering procedure used to compute sparse mNCC. In \name, the camera pose of a real X-ray is estimated using a CNN. Along with the regressed pose, activations at the final convolutional layer are stored. Visualizing these activations shows that the network mostly uses the location of bony structures (\eg spine, pelvis, hips, \etc) to estimate the pose. This activation map is resized to match the original X-ray and used to define a probability distribution over the pixels in the image. For each iteration, a fixed number of patch centers are sampled from this distribution. Specifying the patch size defines the sparse subset of pixels in the detector plane that need to be rendered. In practice, we render 100 patches with a patch size of 13. Since we downsample real X-rays to $256 \times 256$ pixels, when using sparse rendering, we render at most $(100 \cdot 13^2) / 256^2 \approx 25\%$ of the pixels in the image. Note, this is an upper bound because patches can overlap. For ease of visualization, we also render 750 patches with a patch size of 13. Finally, sparse mNCC is computed by averaging the local NCC over all rendered patches and the global NCC computed all rendered pixels. Note that sparse mNCC is a biased estimate of mNCC, and this approach is closely related to prior work that used sparse image patches to estimate mutual information between images \cite{zollei20012d}.

The accuracy and speed of sparse mNCC are provided in \Cref{tab:sparse-mncc-comparisons}. In an alternative formulation of sparse mNCC, we can ignore the distribution defined by the network's activations and instead sample patch centers uniformly at random over the image. We find that this strategy, which we term unbiased sparse mNCC, performs nearly as well as the original sparse mNCC. Finally, we also compare against mNCC. While mNCC was the most accurate image similarity metric and had the highest success rate, it is also the slowest method to compute. On average for one iteration, it is $3.5\times$ faster to render and compute sparse mNCC with 100 patches ($13 \times 13$) than mNCC over all patches. However, since sparse mNCC is a noisy estimate of mNCC computed over all image patches, it takes more iterations to converge than standard mNCC.

\begin{table}[!h]
\centering
\caption{Comparison of sparse mNCC to other mNCC variants.}
\label{tab:sparse-mncc-comparisons}
\begin{tabular}{@{}lcccccc@{}}
\toprule
 & SMSR $\uparrow$ & mTRE (\si{\mm}) $\downarrow$ & Time (\si{\s}) $\downarrow$ \\ \midrule
Sparse mNCC & \successrate & \SI{0.9 \pm 2.8}{} & \SI{2.2 \pm 1.2}{} \\
Unbiased Sparse mNCC & 86\% & \SI{1.0 \pm 2.4}{} & \textbf{\SI{2.1 \pm 1.3}{}} \\
mNCC & \textbf{89\%} & \textbf{\SI{0.8 \pm 2.1}{}} & \SI{3.9 \pm 1.6}{} \\
\bottomrule
\end{tabular}
\end{table}

\section{Converting Imaging System Coordinates to DiffDRR Camera Coordinates}
\label{sec:camera-matrices}
\paragraph{Parsing intrinsic matrices.} 
Intrinsic matrices provided in the DeepFluoro and Ljubljana datasets can be decomposed as
\begin{equation}
    \begin{bmatrix}
        f_x & 0 & x_0' \\
        0 & f_y & y_0' \\
        0 & 0 & 1
    \end{bmatrix} =
    \begin{bmatrix}
        -\sfrac{1}{\Delta X} & 0 & \sfrac{W}{2} \\
        0 & -\sfrac{1}{\Delta Y} & \sfrac{H}{2} \\
        0 & 0 & 1
    \end{bmatrix}
    \begin{bmatrix}
        f & 0 & x_0 \\
        0 & f & y_0 \\
        0 & 0 & 1
    \end{bmatrix} \,,
\end{equation}
where $(f_x, f_y)$ are the focal lengths in the x- and y-directions (in units of pixels), $(x_0', y_0')$ is the camera's principal point (in units of pixels), $(H, W)$ is the height and width of the detector plane (in units of pixels), and $(\Delta X, \Delta Y)$ are the x- and y-direction pixel spacings (in units of length per pixel). From these known parameters, the focal length of the X-ray scanner (in units of length) can be expressed as
\begin{equation}
    f = \frac{f_x \Delta X + f_y \Delta Y}{2} \,,
\end{equation}
and the principal point also expressed in units of length is
\begin{align}
    x_0 &= \Delta X \left( \frac{W}{2} - x_0' \right) \\
    y_0 &= \Delta Y \left( \frac{H}{2} - y_0' \right) \,.
\end{align}
The intrinsic parameters $f$, $\Delta X$, $\Delta Y$, $H$, $W$, $x_0$, and $y_0$ are needed to define the detector plane in DiffDRR~\cite{gopalakrishnan2022fast}.

\paragraph{Parsing extrinsic matrices.} Extrinsic matrices in the DeepFluoro and Ljubljana datasets assume that the camera is initialized at the origin and pointing in the negative z-direction. However, DiffDRR initializes the camera at $(f/2, 0, 0)$ pointed towards the negative x-direction. To transform a camera pose $\mathbf T \in \SE3$ from DeepFluoro/Ljubljana's coordinate system to DiffDRR's coordinate system, we use the following conversion:
\begin{equation}
    \mathbf{\Tilde T} = \mathbf T^{-1}
    \begin{bmatrix}
        0 & 0 & -1 & 0 \\
        0 & 1 & 0 & 0 \\
        1 & 0 & 0 & 0 \\
        0 & 0 & 0 & 1
    \end{bmatrix}
    \begin{bmatrix}
        1 & 0 & 0 & -f/2 \\
        0 & 1 & 0 & 0 \\
        0 & 0 & 1 & 0 \\
        0 & 0 & 0 & 1
    \end{bmatrix} \,.
\end{equation}
When passed to DiffDRR along with the scanner's intrinsic parameters, the camera pose $\mathbf{\Tilde T}$ renders synthetic X-rays using the same geometry as the real-world imaging system (see \Cref{fig:examples}). Transformations of camera poses, along with conversions between multiple parameterizations of $\SO3$ and $\SE3$, are handled using PyTorch3D~\cite{ravi2020pytorch3d}.

\section{Derivation of the Image Formation Model}
\label{app:image-formation-derivation}
For completeness, we present a single, detailed derivation of the X-ray image formation model and preprocessing steps described in many places throughout the main text. Let $\mathbf r(\alpha) = \mathbf s + \alpha (\mathbf t - \mathbf s)$ be a ray originating at the radiation source $\mathbf s \in \mathbb R^3$ and terminating at the target pixel on the detector plane $\mathbf t \in \mathbb R^3$. We are interested in modeling the attenuation of $\mathbf r$ as it travels through the anatomical volume $\mathbf V : \mathbb R^3 \to \mathbb R$. For every point $\mathbf x$ in 3D space, $\mathbf V(\mathbf x)$ returns a \textit{linear attenuation coefficient} that characterizes how much intensity $\mathbf r$ loses to $\mathbf V$ when it travels through $\mathbf x$. A large coefficient denotes that $\mathbf x$ comprises a material that greatly attenuates $\mathbf r$ by absorbing a large amount of its intensity, while a small coefficient represents a material that is easily penetrated. We model points in empty space as having a linear attenuation coefficient of zero. The attenuated intensity of $\mathbf r$ after it has passed through every point on its path, as governed by the Beer-Lambert law~\cite{swinehart1962beer}, is
\begin{equation}
    \label{eq:beer-lambert}
    I(\mathbf p) = I_0 \exp \Big( -\medint\int_{\mathbf x \in \mathbf r} \mathbf V(\mathbf x) \,\mathrm d \mathbf x \Big) \,,
\end{equation}
where $I_0$ is the initial intensity of every X-ray radiating from $\mathbf s$. 

Instead of modeling the attenuated intensity of $\mathbf r$, it is both equivalent and simpler to model the amount of energy absorbed by $\mathbf V$. To this end, we only consider the integral in \cref{eq:beer-lambert} and model the absorption $I_\mu(\mathbf p) = \log I_0 - \log I(\mathbf p)$, which is inversely proportional to $I(\mathbf p)$. If $I_0$ is unknown, we can estimate it by computing the maximum value over all pixels in a set of X-rays acquired using a scanner with the consistent parameters. Then, \cref{eq:beer-lambert} can be expressed as
\begin{align}
    I_\mu(\mathbf p) 
    &= \log I_0 - \log I(\mathbf p) \\
    &= \int_{\mathbf x \in \mathbf r} \mathbf V(\mathbf x) \,\mathrm d \mathbf x \\
    &= \int_0^1 \mathbf V\big(\mathbf r(\alpha)\big) \| \mathbf r'(\alpha)\| \,\mathrm d\alpha \\
    \label{eq:exact}
    &= \| \mathbf p - \mathbf s \| \int_0^1 \mathbf V\big(\mathbf s + \alpha(\mathbf p - \mathbf s)\big) \,\mathrm d\alpha \\
    \label{eq:siddon}
    &\approx \| \mathbf p - \mathbf s \| \sum_{m=1}^{M-1} \mathbf V \left[ \mathbf s + \frac{\alpha_{m+1} + \alpha_m}{2} (\mathbf p - \mathbf s) \right] (\alpha_{m+1} - \alpha_m) \,,
\end{align}
where, in the last step, we approximate $\mathbf V$ with a preprocessed CT volume (a voxel grid of linear attenuation coefficients estimated from the original Hounsfield units). This approximation results in a discretization of the line integral over the voxel grid---note the similarities between \cref{eq:exact} and the \cref{eq:siddon}. The set $\{\alpha_1, \dots, \alpha_M\}$ parameterizes all the intersections of $\mathbf r$ with the orthogonal planes comprising the CT volume, and $\mathbf V[\cdot]$ is an indexing operation that returns the linear attenuation coefficient of the voxel within which a 3D point is located. Siddon's method \cite{siddon1985fast} provides a method for efficiently computing the intersections and indexing operations in the rendering equation~\labelcref{eq:siddon}. In DiffDRR~\cite{gopalakrishnan2022fast}, Siddon's method is reformulated as a series of vectorized tensor operations, enabling \cref{eq:siddon} to be computed in a differentiable manner. Tangentially, one might wonder why the length of the ray $\| \mathbf p - \mathbf s \|$ appears in \cref{eq:siddon}. Note that the SI unit of the linear attenuation coefficient is the reciprocal meter (\si{\per\m}). Therefore, multiplication by term $\|\mathbf p - \mathbf s\|$ serves to make the absorbance $I_\mu(\mathbf p)$ unit-less.

\section{Additional Implementation Details}

\begin{figure}[!h]
    \centering
    \includegraphics[width=0.825\linewidth]{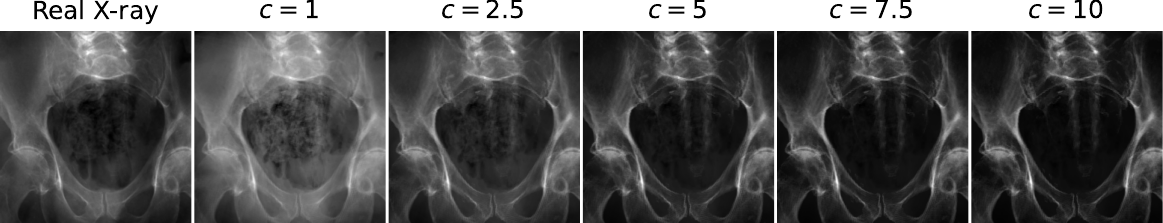}
    \caption{\textbf{Examples of domain randomization via X-ray contrast augmentation.}}
    \label{fig:contrast}
\end{figure}

\paragraph{Domain randomization.} Inspired by previous (non-differentiable) synthetic X-ray renderers~\cite{unberath2018deepdrr}, we augment the contrast of synthetic X-rays by upweighting the attenuation of voxels in the CT scan corresponding to bone. To isolate the voxels corresponding to bone, we segment the CT by thresholding all voxels with Hounsfield units greater than 350. Multiplying these voxels by a bone attenuation multiplier $c \geq 1$ increases the brightness of bones relative to soft tissue. While pretraining the encoder, we randomly sample $c \sim \mathrm{Uniform}[1, 10]$. This domain randomization improves transfer from simulated to real data by diversifying the appearance of synthetic X-rays. Synthetic X-rays rendered with $c \in [1, 10]$ are shown in \Cref{fig:contrast}.

\paragraph{Architecture.} We implemented the pose regression encoder $\mathcal E$ using a ResNet18 backbone. As the rendering and pose regression of synthetic X-rays were performed on the same device, we were limited to a maximum batch size of eight $256 \times 256$ images. This small batch size induced instability in the running estimates of mean and variance in the batch normalization layers, leading us to replace them with group normalization layers. All encoders were trained from scratch for each patient.

\paragraph{Early stopping criteria.} We terminate test-time optimization early if the image similarity metric does not improve by at least 0.05 for 20 iterations in a row.

\paragraph{Hardware.} For each patient, pretraining the pose regression encoder was performed on a single NVIDIA RTX A6000 GPU. For each intraoperative X-ray, test-time optimization was performed a single GeForce RTX 2080 Ti GPU.

\clearpage
\section{Additional Qualitative Results on the DeepFluoro Dataset}
\begin{figure}[!hb]
    \centering
    \includegraphics[width=0.85\linewidth]{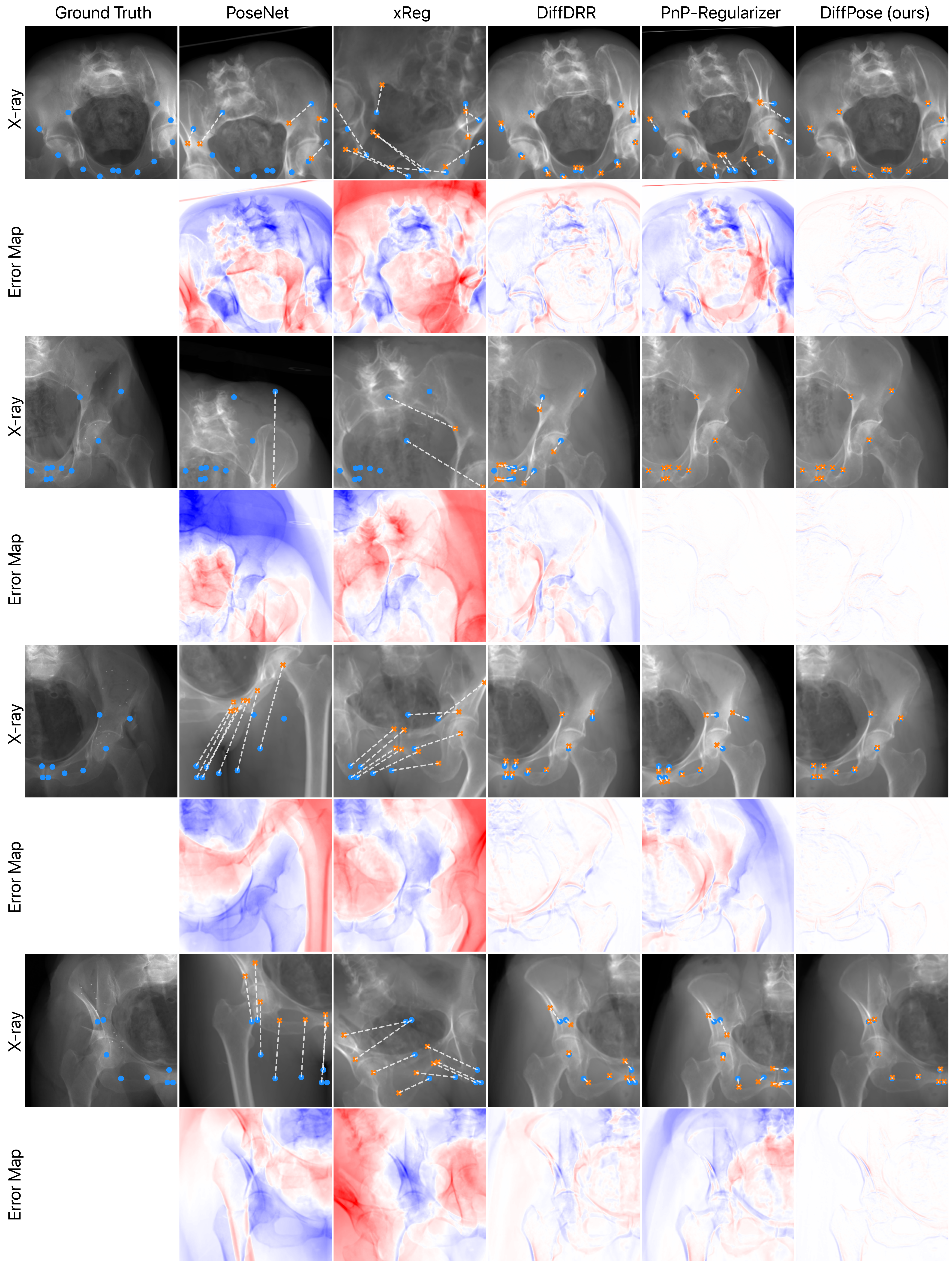}
    \caption{\textbf{Visualizations of additional qualitative results on the DeepFluoro dataset.} Fiducials projected at the ground truth camera pose are in blue, while projections at the estimated pose are in orange. White lines are drawn between corresponding fiducials.}
    \label{fig:deepfluoro-extras}
\end{figure}

\clearpage
\section{Additional Qualitative Results on the Ljubljana Dataset}

The Ljubljana dataset consists of 2D/3D digital subtraction angiography (DSA) images. In 2D DSA images, a subtraction step is used to remove the outline of the skull, attenuating the vasculature and increasing the signal-to-noise ratio \cite{brody1982digital}. In 3D DSA images, the signal-to-noise ratio is too low to capture the smallest blood vessels \cite{schueler20053d}. Despite missing the microvasculature, the trunks of main vasculature are sufficient to drive 2D/3D image registration with \name in most cases.

\begin{figure}[!hb]
    \centering
    \includegraphics[width=\linewidth]{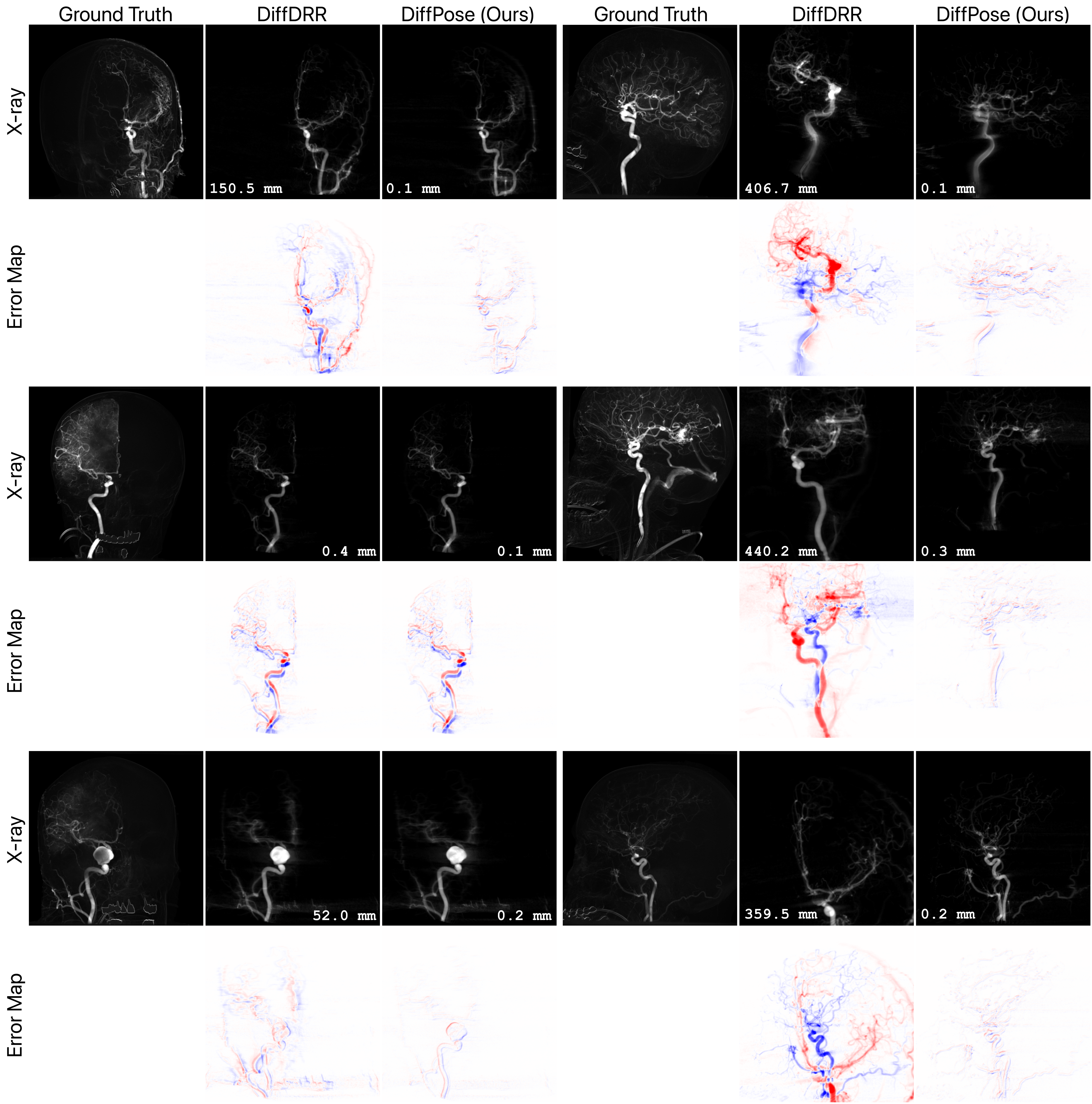}
    \caption{\textbf{Visualizations of additional qualitative results on the Ljubljana dataset.} mTRE is reported for each example.}
    \label{fig:ljubljana-extras}
\end{figure}

\end{document}